\documentclass[letterpaper]{article} 
\usepackage{aaai25}  
\usepackage{times}  
\usepackage{helvet}  
\usepackage{courier}  
\usepackage[hyphens]{url}  
\usepackage{graphicx} 
\usepackage{natbib}  
\usepackage{caption} 
\usepackage{algorithm}
\usepackage{algorithmic}

\usepackage{xspace}
\newcommand{\toolname}{\textsc{LLM Stinger}\xspace}

\usepackage{newfloat}
\usepackage{listings}
\DeclareCaptionStyle{ruled}{labelfont=normalfont,labelsep=colon,strut=off} 
\floatstyle{ruled}
\newfloat{listing}{tb}{lst}{}
\floatname{listing}{Listing}
\title{\toolname: Jailbreaking LLMs using RL fine-tuned LLMs}
\author {
    Piyush Jha,
    Arnav Arora, and
    Vijay Ganesh
}
\affiliations {
    Georgia Institute of Technology, USA\\
    \{piyush.jha, aarora362, vganesh\}@gatech.edu
}

\begin{document}

\maketitle

\begin{abstract}

We introduce \toolname, a novel approach that leverages Large Language Models (LLMs) to automatically generate adversarial suffixes for jailbreak attacks. Unlike traditional methods, which require complex prompt engineering or white-box access, \toolname uses a reinforcement learning (RL) loop to fine-tune an attacker LLM, generating new suffixes based on existing attacks for harmful questions from the HarmBench benchmark. Our method significantly outperforms existing red-teaming approaches (we compared against 15 of the latest methods), achieving a +57.2\% improvement in Attack Success Rate (ASR) on LLaMA2-7B-chat and a +50.3\% ASR increase on Claude 2, both models known for their extensive safety measures. Additionally, we achieved a 94.97\% ASR on GPT-3.5 and 99.4\% on Gemma-2B-it, demonstrating the robustness and adaptability of \toolname across open and closed-source models.


\end{abstract}

%

\section{Introduction}

\begin{table*}[t]
\centering
\resizebox{\textwidth}{!}{%
\begin{tabular}{|c|ccccccccccccccccc|}
\hline
\textbf{\begin{tabular}[c]{@{}c@{}}Attack method / \\ Victim LLM\end{tabular}} &
  \textbf{\begin{tabular}[c]{@{}c@{}}\toolname\\(ours)\end{tabular}} &
  \textbf{GCG} &
  \textbf{GCG-M} &
  \textbf{GCG-T} &
  \textbf{PEZ} &
  \textbf{GBDA} &
  \textbf{UAT} &
  \textbf{AP} &
  \textbf{SFS} &
  \textbf{ZS} &
  \textbf{PAIR} &
  \textbf{TAP} &
  \textbf{TAP-T} &
  \textbf{\begin{tabular}[c]{@{}c@{}}Auto\\DAN\end{tabular}} &
  \textbf{\begin{tabular}[c]{@{}c@{}}PAP\\top5\end{tabular}} &
  \textbf{Human} &
  \textbf{DR} \\ \hline
\textbf{Llama2-7B-chat}     &   \textbf{89.3}   & \underline{32.1} & 19.5 & 15.9 & 0.0  & 0.0  & 3.1  & 19.5 & 3.1  & 0.4  & 6.9  & 5.0  & 3.8  & 0.0  & 0.8  & 0.1  & 0.0  \\
\textbf{Vicuna-7B}          &   \textbf{93.08}   & \underline{89.9} & 83.9 & 83.1 & 17.6 & 16.9 & 18.2 & 76.1 & 52.8 & 26.5 & 66.0 & 67.9 & 78.0 & 89.3 & 15.6 & 46.7 & 20.1 \\
\textbf{Claude 2}           & \textbf{52.2} & -    & -    & 1.5  & -    & -    & -    & -    & -    & 0.6  & \underline{1.9}  & 1.3  & 0.0  & -    & 0.1  & 0.0  & 0.0  \\
\textbf{Claude 2.1}         & \textbf{26.4} & -    & -    & 1.4  & -    & -    & -    & -    & -    & 0.6  & \underline{2.5}  & 1.9  & 0.0  & -    & 0.1  & 0.1  & 0.0  \\
\textbf{GPT 3.5 Turbo 0613} &  \textbf{88.67}  & -    & -    & 44.3 & -    & -    & -    & -    & -    & 20.3 & 52.8 & 54.7 & \underline{78.6} & -    & 10.6 & 25.9 & 16.4 \\
\textbf{GPT 3.5 Turbo 1106} &  \textbf{94.97}  & -    & -    & 56.4 & -    & -    & -    & -    & -    & 33.6 & 42.1 & 45.9 & \underline{60.4} & -    & 11.9 & 3.0  & 36.5 \\
\textbf{GPT 4 Turbo 1106}   &  \underline{80.50}   & -    & -    & 21.4 & -    & -    & -    & -    & -    & 9.3  & 41.5 & 43.4 & \textbf{81.8} & -    & 11.9 & 1.4  & 6.9 \\ 
\hline
\end{tabular}%
}
\caption{Attack Success Rate on HarmBench (standard behaviours test split) for open-source and closed-source victim LLMs. Bold cells highlight the best-performing attack method for each victim LLM, while underlined cells indicate the second best. The dashed (-) cells indicate that the attack method is incompatible with the victim LLM, as it is a closed-source (black-box) LLM.}
\label{tab:results}
\end{table*}

Jailbreaking Large Language Models (LLMs) involve crafting inputs that lead safety-trained models to violate developer-imposed safety measures, producing unintended or harmful responses. One effective method for this is through suffix attacks, where specific strings are appended to the input to trigger undesired behavior. Suffix-based attacks have shown success against both white-box and black-box LLMs, offering a simpler, more efficient, and easily automated alternative without the need for complex prompt engineering and human creativity to craft situations and role-playing templates~\cite{zou2023universal}. Although most of the existing suffix attacks have been patched because of safety training, we observed that modifications of those suffixes can still lead to successful jailbreak attempts. However, manually crafting these modifications or using a white-box gradient-based attacker to find new suffixes is laborious and time-consuming, limiting the scalability of such efforts.

In this work, we introduce \toolname, a tool that uses an LLM to automatically generate highly effective adversarial suffixes that can jailbreak safety-trained LLMs. By fine-tuning an attacker LLM in a reinforcement learning (RL) loop, \toolname generates new attack suffixes for a set of harmful questions from the HarmBench benchmark~\cite{mazeika2024harmbench}. This automated approach efficiently discovers new suffixes that bypass existing defenses, streamlining the process of crafting jailbreak attacks. The RL loop refines the attacker LLM with the help of reward signals that guide it toward more effective attacks. We compared \toolname against 15 SOTA attack methods, and it outperformed all of them on attack success rate. For example, we were able to increase the ASR (+50.3\%) on safety-trained closed-source models such as Claude. 


\begin{figure}[t]
  \centering
  \includegraphics[width=\linewidth]{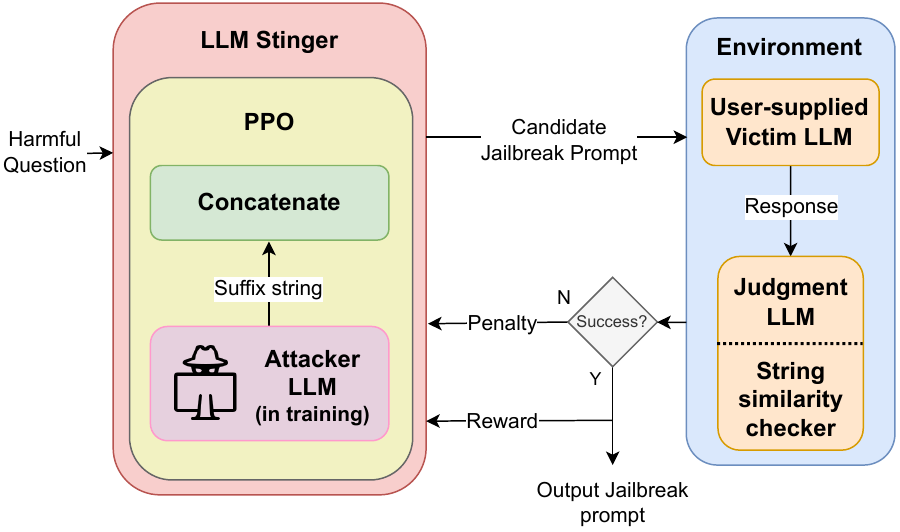}
  \caption{Architecture Diagram of \toolname} 
  \label{fig:arch_diag}
\end{figure}

\section{\toolname}
\textbf{Implementation Details of \toolname.} During training, the attacker LLM takes as input the harmful question and seven publicly available suffixes~\cite{zou2023universal} and is prompted to generate similar new suffixes. The generated suffix is appended to the harmful question (from the training set) and sent to the victim LLM, requiring only black-box access to the model. The victim’s response is evaluated by the judgment model, which provides binary feedback on whether the attack succeeded. Additionally, if the attack fails, a string similarity checker provides token-level feedback, allowing for more precise adjustments to the generated suffixes. This mechanism evaluates how closely the generated suffix aligns with previously successful suffixes, penalizing deviations that stray too far from patterns known to bypass safety measures. This feedback is used to fine-tune the attacker LLM in an RL loop for 50 epochs using the Proximal Policy Optimization (PPO) algorithm, with the goal of improving the attack success rate over multiple iterations (Figure~\ref{fig:arch_diag}). Successful suffixes are saved, and during evaluation on test set questions (disjoint from training set), we assess whether the attacker LLM can generate a question-specific harmful suffix or if any of the 38 newly generated suffixes can successfully trigger an attack.

The string similarity checker plays a critical role in \toolname's adversarial suffix generation process by providing fine-grained (dense) feedback at the token level. Its primary function is to evaluate the degree of similarity between the generated suffix and previously successful suffixes present in the prompt, guiding the attacker LLM to retain essential characteristics that have proven effective in bypassing safety mechanisms. By providing token-level feedback rather than relying solely on binary success or failure signals, the string similarity checker improves the refinement process of suffix generation. This enables more precise adjustments, speeding up finding suffixes that effectively balance new attacks with resemblance to previously successful attacks. Additionally, this checker helps reduce the search space by pruning less promising candidates, making the search process more efficient and improving the overall success rate of generating adversarial suffixes.

\subsection{Implementation and Hardware Details}

The algorithm is implemented using a customized version of the Transformer Reinforcement Learning (TRL) library~\cite{vonwerra2022trl}, which is a comprehensive framework integrated with Huggingface transformers~\cite{wolf-etal-2020-transformers} for training transformer-based language models using reinforcement learning (RL). We employ the Proximal Policy Optimization (PPO) algorithm, frequently used in Reinforcement Learning from Human Feedback (RLHF)~\cite{yang2024harnessing,ouyang2022training,stiennon2020learning}, to fine-tune the language model. Since the original TRL library only supports scalar reward signals for RL fine-tuning, we modified it to handle vector reward signals, enabling fine-grained feedback at the token level. All experiments were performed on a high-performance CentOS V7 cluster with Intel E5-2683 v4 Broadwell processors (2.10 GHz), 2 NVIDIA V100 GPUs, and 64 GiB of memory. 

\section{Experimental Setup} 

We use the standard behavior benchmark from HarmBench that includes a train-test split, and the main metric for evaluation is the attack success rate, a widely used standard in adversarial attack research~\cite{mazeika2024harmbench}. For the attacker model, we chose Gemma, an open-source LLM that allows for system prompt modifications and is not overly cautious in its responses, making it suitable for adversarial purposes. The judgment model is the HarmBench judgment LLM, selected because it outperforms other judgment models (GPT-4, Llama-Guard, AdvBench) on their manually labeled validation sets. To compare \toolname with existing methods, we evaluate it against 15 popular red-teaming approaches and Direct Request (DR) baseline also used by HarmBench. Due to computational limitations, we select a subset of HarmBench victim LLMs. For open-source models, we focus on LLaMA-2, known for its robustness to GCG and other attacks, as well as Vicuna. In the closed-source category, we test on GPT and Claude models, which have undergone extensive safety training with strong defenses at the system and model levels. Additional details are given below.

\subsection{HarmBench Benchmark}

HarmBench is a dataset of harmful behaviors and an evaluation pipeline meant to serve as a general evaluation framework for LLM red-teaming efforts. We use standard behavior benchmarks from HarmBench, which cover a broad range of harms that public LLMs are typically aligned to avoid, such as those against OpenAI, Anthropic, and Meta policies. These are modeled after existing datasets, such as AdvBench and the TDC 2023 Red Teaming Track Dataset~\cite{zou2023universal,mazeika2024harmbench}.

\subsection{Metric}

Attack Success Rate (ASR) is a metric used to evaluate the effectiveness of an attack method on a victim model. It represents the percentage of harmful questions in the test set (post-modification) that are able to manipulate the model into producing unintended or harmful responses or bypassing safety measures leading to a successful jailbreak attack~\cite{zou2023universal,chao2023jailbreaking,mehrotra2023tree}.

\subsection{Attack Methods}

Here is a list of the various attack methods we use for comparison:

\begin{itemize}
    \item GCG \cite{zou2023universal} optimizes tokens of adversarial suffix to maximize log probability of affirmative string being generated.
    \item GCG-Multi \cite{zou2023universal} optimizes a single suffix for multiple user prompts (for a single LLM).
    \item GCG-Transfer \cite{zou2023universal} extends GCG-Multi by optimizing across multiple open-source models, but applies the same attack suffixes discovered from these open-source models to closed-source models without further modification.
    \item PEZ \cite{wen2024hard} optimizes tokens of adversarial suffix through nearest-neighbor projection. 
    \item GBDA \cite{guo2021gradient} uses Gumbel-softmax distribution to optimize tokens of adversarial suffix.
    \item UAT \cite{wallace2019universal} uses Taylor approximation of first order around token embedding’s gradient to optimize tokens of adversarial suffix.
    \item AutoPrompt (AP) \cite{shin2020autoprompt} is a technique similar to GCG, but uses a different candidate selection strategy.
    \item Stochastic Few-Shot (SFS) \cite{perez2022red} creates test cases through few-shot generation with attacker LLM.
    \item Zero-Shot (ZS) \cite{perez2022red} creates test cases through zero-shot generation with attacker LLM.
    \item PAIR \cite{chao2023jailbreaking} elicits harmful behavior through prompting and exploration from attacker LLM.
    \item TAP \cite{mehrotra2023tree} is a technique similar to PAIR, but uses tree-structured prompting.
    \item TAP-Transfer \cite{mehrotra2023tree} builds on TAP by transferring to other LLMs.
    \item AutoDAN \cite{liu2023autodan} starts with handcrafted prompts and uses hierarchical genetic algorithms to modify these prompts.
    \item PAP \cite{zeng2024johnny} employs persuasive strategies to make the attacker LLM sound more convincing with each request.
    \item Human Jailbreaks \cite{shen2023anything} are a set of handcrafted jailbreaks.
    \item Direct Requests serves as a baseline where the harmful request is provided with no modifications.
\end{itemize}

\subsection{Victim LLMs}

Due to computational limitations, we selected a subset of victim LLMs used by HarmBench. We selected Llama 2~\cite{touvron2023llama}, which has demonstrated strong resilience against numerous attacks, including GCG. We also included Vicuna~\cite{chiang2023vicuna}, as it is fine-tuned from Llama. Among closed-source models, we chose GPT-3.5 and GPT-4 APIs~\cite{achiam2023gpt}, along with the Claude APIs~\cite{bai2022constitutional}, both of which incorporate defenses at both the system and model levels.

\section{Results}

Our method demonstrates a significant improvement in jailbreaking LLMs, particularly those that have undergone safety training, where other attack methods have struggled. As shown in Table~\ref{tab:results}, on Llama2-7B-chat, we observed the highest increase in Attack Success Rate (ASR), with our method achieving a +57.2\% improvement compared to the next best approach. This substantial increase highlights the efficacy of our technique in bypassing the defenses of highly safety-trained models. 

When evaluating closed-source (black-box) models, our method increased the ASR by +50.3\% on Claude 2, a model known for its rigorous safety training. In comparison, the next best method only reached a 1.9\% ASR, further highlighting the strength of our approach. On GPT-3.5, we achieved the highest ASR of 94.97\%, showcasing the robustness of our method across different closed-source models. A key advantage of our approach is that it only requires black-box access to the victim model, unlike many of the attack methods in HarmBench, which need white-box access. This allows our method to support a wider range of models, as it doesn’t rely on internal model weights, making it more versatile and applicable to both open and closed-source LLMs. We also tested our method on Gemma-2B-it, the same model we use as our attacker LLM, which is not included in the HarmBench benchmark. Remarkably, we achieved an astounding ASR of 99.4\% on this model using our method, further demonstrating its adaptability and effectiveness across different models. Since we use a judgment LLM classifier model to assess the success of each attack during feedback, we took extra steps to manually verify the outputs from the victim models. This ensures that our approach is genuinely breaking through the model's defences, rather than gaming the classifier.

\section{Related Work}

Research on adversarial attacks against large language models (LLMs) has progressed rapidly, evolving from manually crafted prompts and white-box attacks to fully automated and scalable methods. As LLM deployment has expanded, so has the need to systematically understand, generate, and defend against jailbreak attacks. Prior work spans suffix-based optimization, role-playing and conversational attacks, evolutionary and fuzzing-based methods, large-scale evaluations, and emerging defense strategies. We organize the related work into thematic categories below.

\subsection{Suffix-Based and Optimization-Driven Attacks}

\citet{zou2023universal} introduce universal adversarial suffixes optimized using greedy and gradient-based search techniques, demonstrating strong transferability across models. \citet{geiping2024coercing} generalize optimization-based attacks beyond alignment bypassing to include prompt extraction, misdirection, denial of service, and control flow manipulation, showing that safety training alone is insufficient to prevent coercive behaviors. \citet{pasquini2024neural} propose Neural Exec, which uses optimization-driven prompt injection triggers that can bypass preprocessing stages in Retrieval-Augmented Generation pipelines and evade blacklist-based defenses.

\subsection{Role-Playing and Conversational Jailbreak Attacks}

\citet{tian2023evil} propose Evil Geniuses, a Red-Blue team inspired role-playing framework that autonomously generates aggressive jailbreak prompts using GPT-3.5. \citet{liu2023autodan} introduce AutoDAN, which employs a hierarchical genetic algorithm with selection, crossover, and mutation operators to iteratively refine role-playing jailbreak prompts based on a fitness score. \citet{chao2023jailbreaking} propose Prompt Automatic Iterative Refinement (PAIR), which uses an attacker LLM to generate role-playing prompts and a judgment LLM to provide feedback, forming a fully automated refinement loop.

\subsection{Tree-Based, Evolutionary, and Fuzzing Approaches}

\citet{mehrotra2023tree} propose Tree of Attacks with Pruning (TAP), which uses tree-of-thought reasoning to iteratively refine jailbreak prompts under black-box access constraints, guided by an evaluator LLM. \citet{yao2024fuzzllm} introduce FuzzLLM, a universal fuzzing framework that combines jailbreak constraints with prohibited queries to systematically identify vulnerabilities across multiple LLMs. \citet{chu2024comprehensive} provide a comprehensive empirical evaluation of 13 jailbreak methods across multiple models and violation categories, demonstrating that optimized and automated prompts consistently achieve high attack success rates.

\subsection{Evaluation and Measurement of Jailbreak Vulnerabilities}

\citet{wang2023decodingtrust} analyze GPT-3.5 and GPT-4 across multiple trust dimensions, observing that stronger models may be more susceptible to jailbreaking despite improved benchmark performance. \citet{wei2024jailbroken} conduct a large-scale evaluation of suffix-based and role-playing jailbreaks, arguing that jailbreak vulnerabilities may be inherent to current safety training paradigms and are not resolved through model scaling alone. \citet{sun2024trustllm} introduce TrustLLM, a benchmark spanning multiple trust dimensions and models, finding correlations between trustworthiness and utility. \citet{xu2024llm} systematically compare attack and defense methods across several LLMs, highlighting the importance of universal attacks and the role of special tokens in jailbreak success.

\subsection{Defense Strategies Against Jailbreak Attacks}

\citet{zhou2024defending} propose the InContext Adversarial Game (ICAG), which formulates defense as an iterative adversarial interaction between attacker and defender agents without requiring fine-tuning. This agent-based framework enables adaptive defenses against evolving jailbreak strategies.

\subsection{Learning-Based Attack Generation and Reinforcement Learning}

Most prior jailbreak attacks either rely on white-box gradient-based optimization, prompt-level manipulations such as role-playing, or manually identified jailbreak patterns. In these approaches, the attacker model itself is not fine-tuned. In contrast, we are the first to fine-tune an attacker LLM to generate new and modified versions of suffix-based attacks. Our feedback-driven framework enables automated attack evolution, introducing a new paradigm for adversarial LLM research.

The use of language models within reinforcement learning loops for vulnerability discovery has been explored in other domains. \citet{jha2023bertrlfuzzer} combine a BERT-based language model with reinforcement learning and binary feedback to generate grammar-adhering mutations for web vulnerability discovery. Our approach differs by leveraging a large language model to generate complete attack vectors and by using fine-grained symbolic feedback signals, drawing inspiration from Reinforcement Learning with Symbolic Feedback \citep{jha2024rlsf}.

\section{Conclusions}

We present LLM Stinger~\cite{jha2025llm}, a new adversarial attack framework designed to jailbreak safety-trained LLMs LLMs by generating adversarial suffixes within an RL loop. Tested against 15 attack methods across 7 LLMs, LLM Stinger showed a 52.2\% attack success rate (ASR) on Claude 2, a model specifically designed to prevent such attacks, compared to the next best method’s 1.9\% ASR. The system's scalability and efficacy demonstrate its capability to break through various LLMs using only black-box API access. 

It is easy to see that the jailbreak attack search problem addressed here is combinatorially hard, and hence naive methods won't work. Our RL-based heuristic search technique is effective as it is guided by feedback from a judgment LLM and a string similarity checker that helps prune the search space dramatically and focus its exploration on high-potential regions. As a result, it is not only able to discover new attack suffixes efficiently but also adapts to the defenses of highly safety-trained models, thus demonstrating the practical strength of our approach. In the future, we plan to include support for additional token-level attack methods, conduct comparisons against more attack strategies and victim LLMs, including multi-modal models, and explore additional feedback mechanisms to enhance the effectiveness of the attacker LLMs.



\bibliography{main}

\end{document}